\documentclass{article}

\usepackage[final]{nips_2016}

\usepackage[utf8]{inputenc} 
\usepackage[T1]{fontenc}    
\usepackage{hyperref}       
\usepackage{url}            
\usepackage{booktabs}       
\usepackage{amsfonts}       
\usepackage{nicefrac}       
\usepackage{microtype}      

\usepackage{amsmath} 
\usepackage{graphicx}
\usepackage{algorithm,algorithmic}
\floatname{algorithm}{Procedure}

\newcommand{\R}{\mathbb{R}}
\newcommand{\Z}{\mathcal{Z}}
\newcommand{\X}{\mathcal{X}}

\title{Latent Geometry and Memorization in Generative Models}

\author{
  Matt D.~Feiszli \\ 
  Sentient Technologies\\
  San Francisco, CA 94111 \\
  \texttt{matt.feiszli@sentient.ai} \\
}

\begin{document}

\maketitle

\begin{abstract}
It can be difficult to tell whether a trained generative model has learned to generate novel examples or has simply memorized a specific set of outputs.  In published work, it is common to attempt to address this visually, for example by displaying a generated example and its nearest neighbor(s) in the training set (in, for example, the $L^2$ metric).  As any generative model induces a probability density on its output domain, we propose studying this density directly.  We first study the geometry of the latent representation and generator, relate this to the output density, and then develop techniques to compute and inspect the output density.  As an application, we demonstrate that "memorization" tends to a density made of delta functions concentrated on the memorized examples.  We note that without first understanding the geometry, the measurement would be essentially impossible to make.

\end{abstract}

\section{Introduction}
\label{section_intro}

Variational Auto-Encoder (VAE) and Generative Adversarial Network (GAN) models have enjoyed considerable success recently in generating natural-looking images.  However, in many cases it can be difficult to tell when a trained generator has actually learned to generate new examples; it is entirely possible for a VAE to simply memorize a training set.  GAN training provides only indirect access to a training set, so direct memorization is less of an issue.  However, it is still possible for any generative model to concentrate its probability mass on a small set of outputs, and the intrinsic dimension of the output is unclear.  To identify memorization, experimenters often provide visual evidence: e.g. some generated examples may be shown alongside their nearest neighbors in a training set (e.g. [1], [2]).  If the neighbors differ from the generated example, memorization is declared to be unlikely.  Alternatively, outputs along a path in latent space may be plotted; if the output changes smoothly this suggests generalization, whereas sudden changes suggests memorization.

Instead, we propose studying the induced probability density on output space.  To do this, we must change variables and transform a density on the latent space to a density on the output space.  This computation requires some care, as the data lie on low-dimensional submanifolds of the input and output space, and the standard formulas will become degenerate and fail.  In what follows we first establish the local geometry of the situation, which allows us to obtain a formula for the output density.  We then introduce ways to measure the degree to which a generator has memorized, and show experimental results.

This enables us to characterize "memorizing" as learning a probability density on output space which concentrates its mass on a finite number of points (in the limit, the learned measure tends to a collection of delta functions).  In contrast, generalization implies a density which smoothly interpolates points, assigning mass to large regions of output space.

\section{Mapping Latent Space to Output Space}
\label{section-mapping}

Consider a trained generative model, where a learned (but now fixed) generator function $f$ maps a space of random variables $Z \subseteq \R^m$ to an output space $X \subseteq \R^n$.  We assume our generator mapping $f:\R^m \to \R^n$ is differentiable.  (This may be false, particularly when $f$ is a neural network with non-differentiable nonlinearities, but $f$ will still be piecewise smooth.)  We further assume that $f^{-1}$ is so difficult to compute that it is effectively unavailable.

A main difficulty is that the latent space $\R^m$ may be large relative to the intrinsic dimension of the learned representation $\Z$.  $m$ is typically chosen to be "large enough" for the problem at hand, and may be larger than necessary.  That is, the learned latent representation may have dimension $l < m$.  Assuming the typical case where $m << n$, we observe that $f$ can only map the latent space onto a submanifold of the output space with dimension at most $l \le m < n$.  Thus we see that $\Z \subset \R^m$ is a latent (sub)manifold of dimension $l$ and $f(\Z) = \X \subset \R^n$ is our output (sub)manifold of dimension $\le l$.

In particular, as a map $\R^m \to \R^n$, we see that $f$ is degenerate as its range lies on a low-dimensional submanifold.

\subsection{Tangent Spaces, Singular Vectors, and the Volume Element}

While global understanding of $f:\R^m \to \R^n$ is not possible in general, the local behavior can be understood by computing the Jacobian matrix $J_f$ and considering the linearized map
\begin{align*}
    f(z + h v) = f(z) + h J_f v + o(h)
\end{align*}
for $h \in \R$ small.  In particular, the rank of $J_f$ tells us about the intrisic dimension of the manifold near a point (see [3]).  Further, the singular value decomposition (SVD) allows us to write
\begin{align*}
    J_f = U \Sigma V^T
\end{align*}
where $V, U$ are orthogonal matrices whose columns (the "singular vectors") span $\R^m$ and $\R^n$ respectively, and $\Sigma$ is a diagonal matrix (the "singular values").  It follows that the right and left singular vectors corresponding to non-zero singular values form a basis for the tangent spaces to $\Z$ and $\X$ ([3], [4]).  The singular vectors $v_i$ of $V$ with degenerate (i.e. $0$) singular values correspond to subspaces which get collapsed (via projection) onto the tangent space before the linearized mapping $f$.

In other words, if $J_f$ has $l$ non-zero singular values, $f$ locally maps $\R^m$ onto an $l$-manifold in $\R^n$.  Moving in directions $v_i \in \R^m$ corresponding to large singular values $\sigma_i$ will cause greater change (in the $L^2$ distance) in the output than directions with smaller $\sigma_i$.  (See section \ref{section-intrinsic-dimension} for experimental analysis of the intrinsic dimension.)

Assuming we have $l < m < n$ non-zero singular values at some point $Z$, then (as mentioned above) $f$ is degenerate and it does not make sense to talk about a volume element.  However, we can consider the restriction of $f$ as a map $\Z \to \X$ between $l$-manifolds.  This restriction of $f$ will still be a diffeomorphism from $\Z$ onto its image, and we can talk about a volume element here.  In particular, the change-of-variable formula will give 
\begin{align}
\label{eqn-volume-element}
    dVol_\X = \left( \prod_{\sigma_i \ne 0} \sigma_i \right) dVol_\Z
\end{align}
That is, volumes on $\X$ and $\Z$ differ by the product of the nonzero singular values at corresponding points $X = f(Z)$.
\section{The Density on Output Space}
\label{section_mapping}

The random vectors $Z = (z_1, z_2, ... z_m)$ are typically drawn from distributions that are easy to sample, e.g. each $z_k$ may be an independent normal or uniform random variable.  Whatever the distribution of $p(Z)$ is, in conjunction with $f$ it induces a density $\widetilde p$ on outputs $X = f(Z)$; in the case $m = n$ the induced density would have (by change-of-variable) the well-known form
\begin{align*}
    \widetilde p(X) = p(Z) \frac{dVol(\Z)}{dVol(\X)} = \frac{p(Z)}{|J_f|}
\end{align*}
where $J_f$ is the Jacobian matrix of $f$ (implicitly at $Z$), and $|\cdot|$ denotes the determinant (recall the determinant describes the volume element and is the product of eigenvalues of $J_f$).  However, we have $m < n$ and cannot use this formula directly.  If $J_f$ had rank $m$ we could replace the denominator with $\sqrt{|J_f^TJ_f|}$, but as discussed above, we find ourselves in a still more degenerate case.

However, using equation \ref{eqn-volume-element} we can compute the induced density as 
\begin{align}
\label{eqn-induced-density}
    \widetilde p(X) = p(Z) \frac{dVol(\Z)}{dVol(\X)} = \frac{p(Z)}{\prod_{\sigma_i \ne 0} \sigma_i}
\end{align}
where $\{\sigma_i \}_{i=1}^l$ are the singular values of $J_f$ at $Z$.  Note that we can also restrict to even lower-dimensional problems by discarding more singular values and vectors; in practice one typically sets a threshold below which any singular values are considered to be zero.

\section{Measuring Memorization}
\label{section-measuring}
We introduce two measurements: the first is based on the density $\widetilde p(X)$ on lines (in latent space) joining two outputs, and considers how much the density drops in-between sample points.  The second measure is based on the local rate of decay of the density about individual points, and provides a local measure of the density's concentration.

\begin{figure}[ht]
    \begin{center}
    \includegraphics[scale=.5,trim=100 100 100 60, clip]{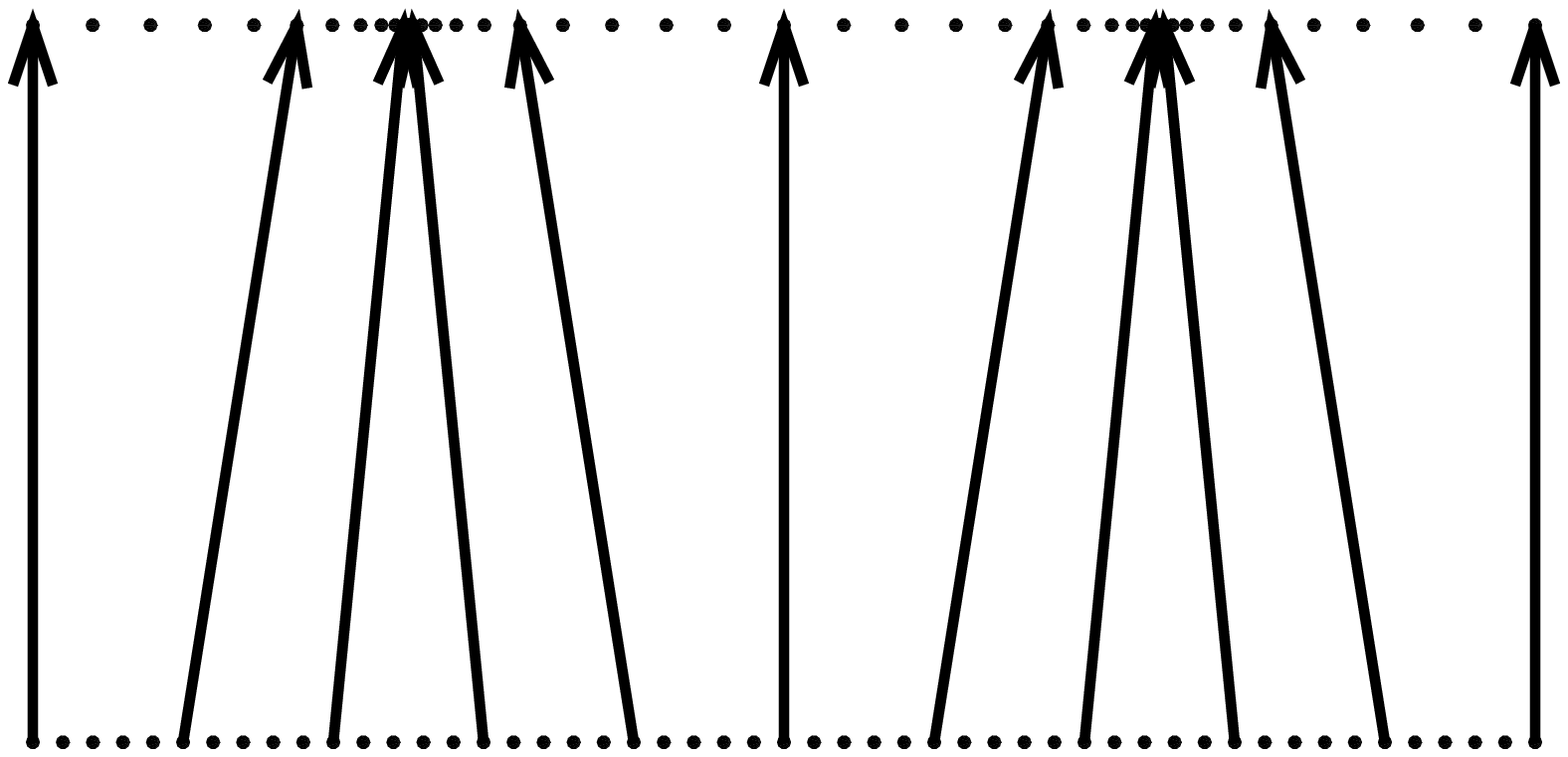}
    \end{center}
    
    \caption{A mapping $f$ taking a uniform density in $\Z$ to a density $\widetilde p(X)$ concentrated on two points in $\X$.  Dot spacing corresponds to density, i.e. $\widetilde p(X)$ is less dense in-between the points of concentration in $\X$. }
    
    \label{fig-density-arrows}
\end{figure}

It may help to consider figure \ref{fig-density-arrows}:  if $\widetilde p(X)$ has memorized a few examples, it implies $f$ must map large regions of $\Z$ to small regions of $\X$.  If we plot $\widetilde p(X)$ as a function of distance (in $\X$), we can examine both the drop in density between output samples and the rate at which the density decays around a given example or examples.  Particularly for considering the decay rate, it is important that we consider $\widetilde p(X)$ as opposed to $\widetilde p(f(Z))$ -- that is, we should use distances on $\X$, not distances in $\Z$.  In cases of memorization, large regions in $\Z$ correspond to nearly constant output.  This implies the density $\widetilde p$ will appear to be spread over "large" neighborhoods when viewed as a function on $\Z$, while simultaneously appearing obviously concentrated in $\X$ (see again figure \ref{fig-density-arrows}).

\subsection{The Density Along Lines}
\label{section-lines}

When latent representations $z_1, z_2$ for outputs $x_1, x_2$ are known, one can construct a path joining the sample points in latent space.  For example, the linear path
\begin{align*}
\gamma_z(t) \equiv (1 - t)z_1 + t z_2 \qquad , t \in [0, 1]
\end{align*}
corresponds immediately to a path in output space 
\begin{align*}
    \gamma_x(t) \equiv f(\gamma_z(t))
\end{align*}
However, we wish to detect point masses in output space, which means we should measure distances there as well.  This unfortunately means we need to choose a metric on $\X$.  For a VAE, the standard reconstruction loss obtained from assuming a Gaussian distribution on output space leads to a square loss on output space.  This means the $L^2$ metric is in some sense natural in this case, and it is what we will use.\footnote{The $L^2$ metric is typically a poor metric for comparing, say, images.  However in our case a metric which made related objects seem close, independent of their visual details, might actually be a disadvantage.  Using a metric in which all objects of a certain type were at distance zero from each other, we could not tell if the generator had memorized or not.}  We can then reparametrize to obtain $\gamma_x(s)$ by constructing $s(t)$ as
\begin{align*}
    s(t) = \int_0^t \left\| \frac{\partial \gamma_x}{\partial t} (\eta) \right\|_{L^2} d\eta
\end{align*}
and then inverting the mapping (numerically) to obtain $t(s)$.  Finally, we can plot $\widetilde p(s)$.  In practice we compute the integral $s(t)$ by summing $L^2$ distances between sample points along a discrete curve as described in procedure \ref{proc-along-lines}.

In summary:
\begin{algorithm}
\caption{Computing density along paths in $\X$}
\label{proc-along-lines}

\begin{enumerate}
\item Consider a path $\gamma_z(t)$ joining two latent points and sample it at times $\{ t_k \}_{k=1}^N$.  Let $z_k \equiv \gamma_z(t_k)$.
\item Using $f$, construct a discrete path $x_k = f(z_k)$ in $\X$.
\item Set $s_1 = 0$ and $s_k = \sum_{j=1}^k \| x_{k+1} - x_k \|_{L^2}$
\item Using equation \ref{eqn-induced-density} this yields the discrete function 
\begin{align*}
\widetilde p(s_k) = \frac{p(t_k)}{\prod_{\sigma_i \ne 0}\sigma_i}  \qquad , k = 1, 2, ..., N    
\end{align*}
where the $\sigma_i$ are singular values of $J_f$ at $z_k$.
\end{enumerate}
\end{algorithm}

Remarks:
\begin{enumerate}
\item When looking along paths joining two endpoints, it is possible for the path $\gamma_z$ to come near the latent representation of a third point.  In this case, the density may noticeably rise in the middle, or it may simply cause a falsely high density in between the endpoints.
\item To avoid the false readings above, we suggest either
\begin{enumerate}
\item Using the local measure described next, or
\item Computing only along paths joining nearest neighbors.  Indeed, perhaps the most convincing evidence of memorization comes from computing  paths joining two nearby instances of a single class and seeing essentially no probability mass in the middle.
\end{enumerate}
\end{enumerate}




\subsection{Local Measures}
\label{section-decay}

Rather than consider paths between sample points, one might wish for a local measure of concentration.  One alternative to interpolating between endpoints would be to compute the probability mass contained in balls of radius $\epsilon$ (measured in $\X$, not $\Z$) about a given set of points.  If the mass increases rapidly as a function of $\epsilon$, this indicates memorization.  However, this is essentially noncomputable, as we'd need to integrate the density over a very high-dimensional ball, and evaluating the density requires $f^{-1}$, which is unavailable.  

It might be possible to use some random sampling or other methods to overcome or mitigate the objections above.  However, looking at the rate of decay of $\widetilde p(X)$ along a collection of paths passing through a point seems to provide a good measure of concentration.  We consider a set of lines in $\Z$ passing through a given point $Z$ and apply similar methodology to procedure \ref{proc-along-lines}.  The question is which lines are informative?

Note that choosing random directions or, say, every coordinate direction, will generally not work well.  Section \ref{section-mapping} explains why: one should consider decay only in nondegenerate singular directions.  Using degenerate singular directions amounts to measuring the density along paths on which the output is constant, or nearly so.  This does nothing but add noise and numerical instability to the calculations, and in the worst case (which can be easily observed by choosing degenerate directions, see figure \ref{fig-degenerate-direction}) renders memorization and generalization indistinguishable.

Similar to procedure \ref{proc-along-lines}, we have procedure \ref{proc-decay} for computing decay:
\begin{algorithm}[ht]
\caption{Computing decay of $\widetilde p$ about points in $\X$}
\label{proc-decay}

\begin{enumerate}
\item Consider a path $\gamma_z(t) = z_0 + t v$ for a nondegenerate singular vector $v$ and point $z_0$.  
\item Sample $\gamma_z$ at times $\{ t_k \}_{k=-N}^N$.  Let $z_k \equiv \gamma_z(t_k)$ (so $z_0$ here agrees with $z_0$ above). 
\item Using $f$, construct $x_k = f(z_k)$ in $\X$.
\item Set $s_0 = 0$ and $s_k = \sum_{j=1}^k \| x_{k+1} - x_k \|_{L^2}$ for $k > 0$ (similar for negative $k$, but distances are negative).
\item Using equation \ref{eqn-induced-density} this yields the discrete function 
\begin{align*}
\widetilde p(s_k) = \frac{p(t_k)}{\prod_{\sigma_i \ne 0}\sigma_i}  \qquad , k = -N, -(N-1), ..., 0, ..., N-1, N    
\end{align*}
where the $\sigma_i$ are singular values of $J_f$ at $z_k$.
\end{enumerate}
\end{algorithm}

Remarks:
\begin{itemize}
\item There are several ways to combine the decay measures from each singular direction into a single measure.  In section \ref{section-results} we propose means of second differences of $\log \widetilde p(s)$ (a measure of peakiness or concentration), but other variants (e.g. maximum) are possible.
\end{itemize}

\section{Computation and Results}

Here we discuss some computational details and illustrate the methods of section \ref{section-measuring} on a pair of VAEs trained on the MNIST dataset.  Identical architectures, one VAE was well-trained while the other was trained to overfit and memorize the data set.

\subsection{Computing the Jacobian $J_f$}

Our experiments were performed with the Keras frontend to Tensorflow.  While Tensorflow supports automatic differentiation for scalar-valued functions, there is no support for automatic  differentiation of vector-valued functions (this is awkward to implement using reverse-mode automatic differentiation).  Hence, we are unable to use autodiff to compute Jacobian matrices.  We instead use a simple central-difference approximation for each entry $\{J_f\}_{k,m}$
\begin{align*}
    \frac{\partial f_k}{\partial z_m} \approx \frac{1}{2\epsilon} \left(f_k(z_m + \epsilon) - f_k(z_m - \epsilon) \right)
\end{align*}
for some small $\epsilon$.  This requires $\mathcal{O}(MN)$ function evaluations for a latent space of size $M$ and output space of size $N$.  However, each evaluation is a neural network forward pass and easily parallelizable.

\subsection{The Intrinsic Dimension of the Manifolds}
\label{section-intrinsic-dimension}
\begin{figure}[!htb]
    \begin{center}
    \includegraphics[scale=.25, trim=60 0 60 40, clip]{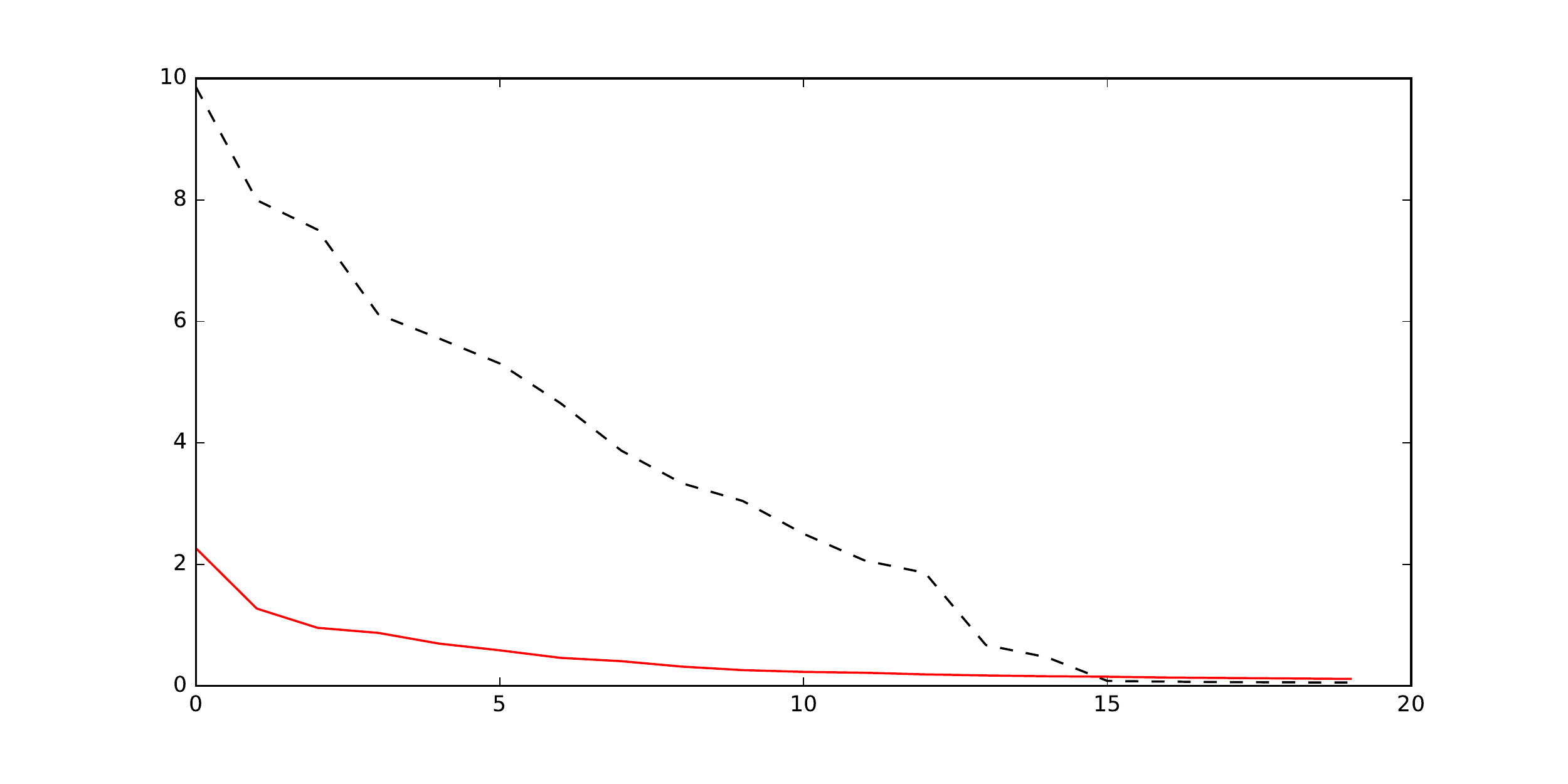} \qquad
    \includegraphics[scale=.25, trim=60 0 60 40, clip]{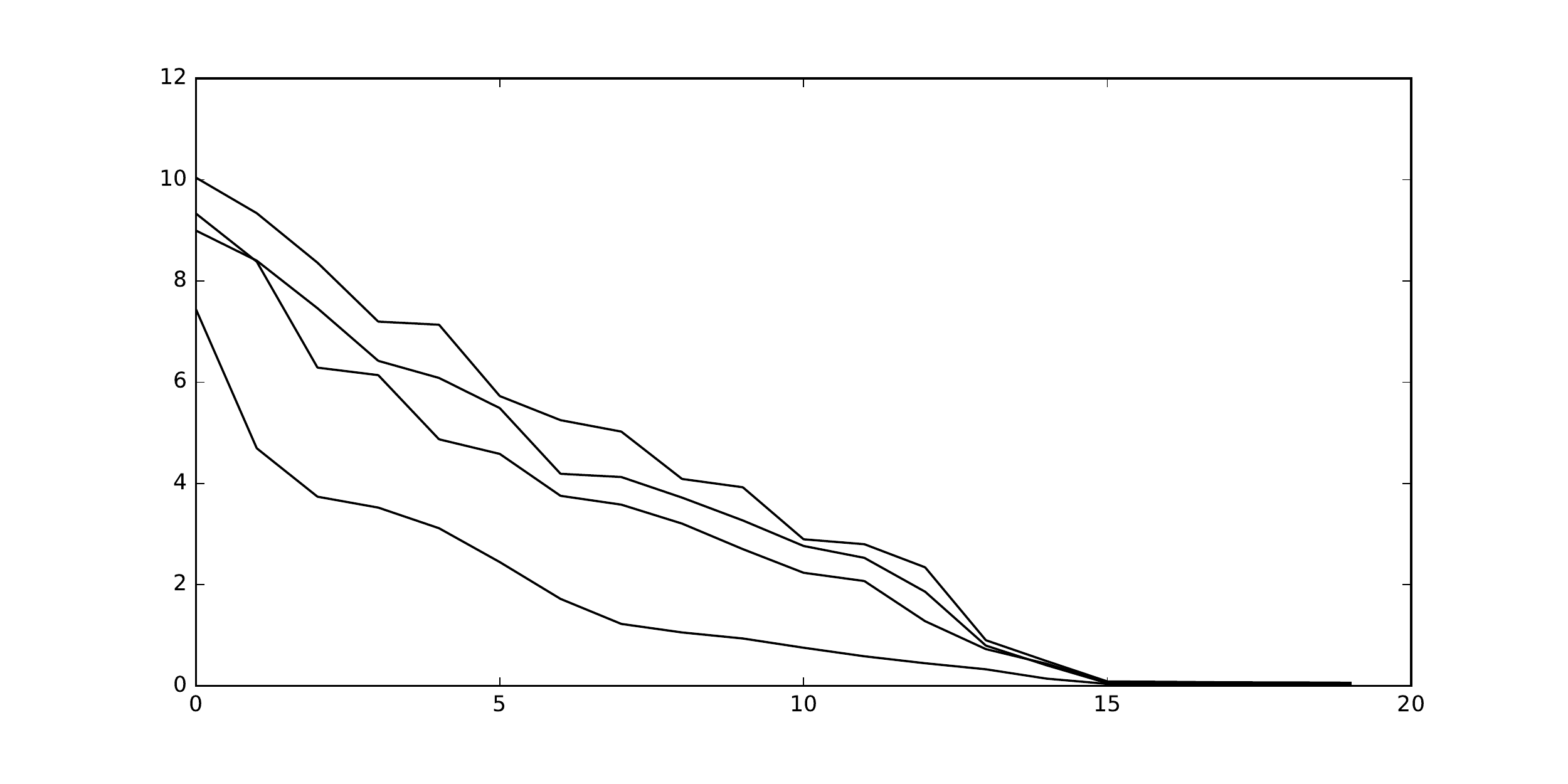}
    \end{center}
    
    \caption{Left: Mean decay of first 20 singular values of the Jacobian at 1000 randomly-chosen training samples suggest an intrisic dimension $\le 14$.  Right: Decay of singular values near four points from different classes.  The effective dimension appears lower near certain examples than others.}
    
    \label{fig-singular-values}
\end{figure}

The SVD of the Jacobian tells us about the intrinsic dimension of the generator function near a point.  In figure \ref{fig-singular-values} we plot the 20 largest singular values in decreasing order, averaged over 1000 randomly-chosen points in the training set.  For both the well-trained and overfit networks, there are no significant singular values beyond the 14th, and particularly for the overfit network the decay begins earlier.  The well-trained network has larger singular values overall, reflecting the fact that its generator covers greater volumes of the output space.  It seems reasonable to declare the intrinsic dimension of either map to be no greater than 14 dimensions.  (Applying SVD to the point cloud of latent representations, as opposed to tangent spaces, also suggests the dimension of the latent representation is no greater than 14.)  In the second plot, we show decay of singular values about 4 training examples from different classes in the well-trained network.  These decay at various rates, suggesting that the effective dimension of the map varies across classes, so locally the map may be lower-dimensional in some areas.

\subsection{Results}
\label{section-results}
We use the methods of section \ref{section-measuring} to explore the differences between two VAEs trained on MNIST.  Each used an identical architecture consisting of several convolutional layers with ReLU and max-pooling, followed by fully-connected layers to compute means and log-variances for the latent distributions, followed by fully-connected layers and transposed convolutions with strides (each with ReLU) for the generator.  Data was normalized to lie in the range $[-1, 1]$ and a final $\tanh$ output layer was applied.

Both models used a 100-dimensional latent representation with a latent prior of independent and $\sim \mathcal{N}(0, 1)$.  However, the first was trained on only 100  examples, while the second trained on the entire 60,000 example training set.  The difference was stark.  The overfit model had huge dips in $\widetilde p(s)$ between training samples, and the decay measure also showed huge concentration of mass on training examples. 

\begin{figure}[!htb]
    \begin{center}
    \includegraphics[scale=.275, trim=80 40 80 40, clip]{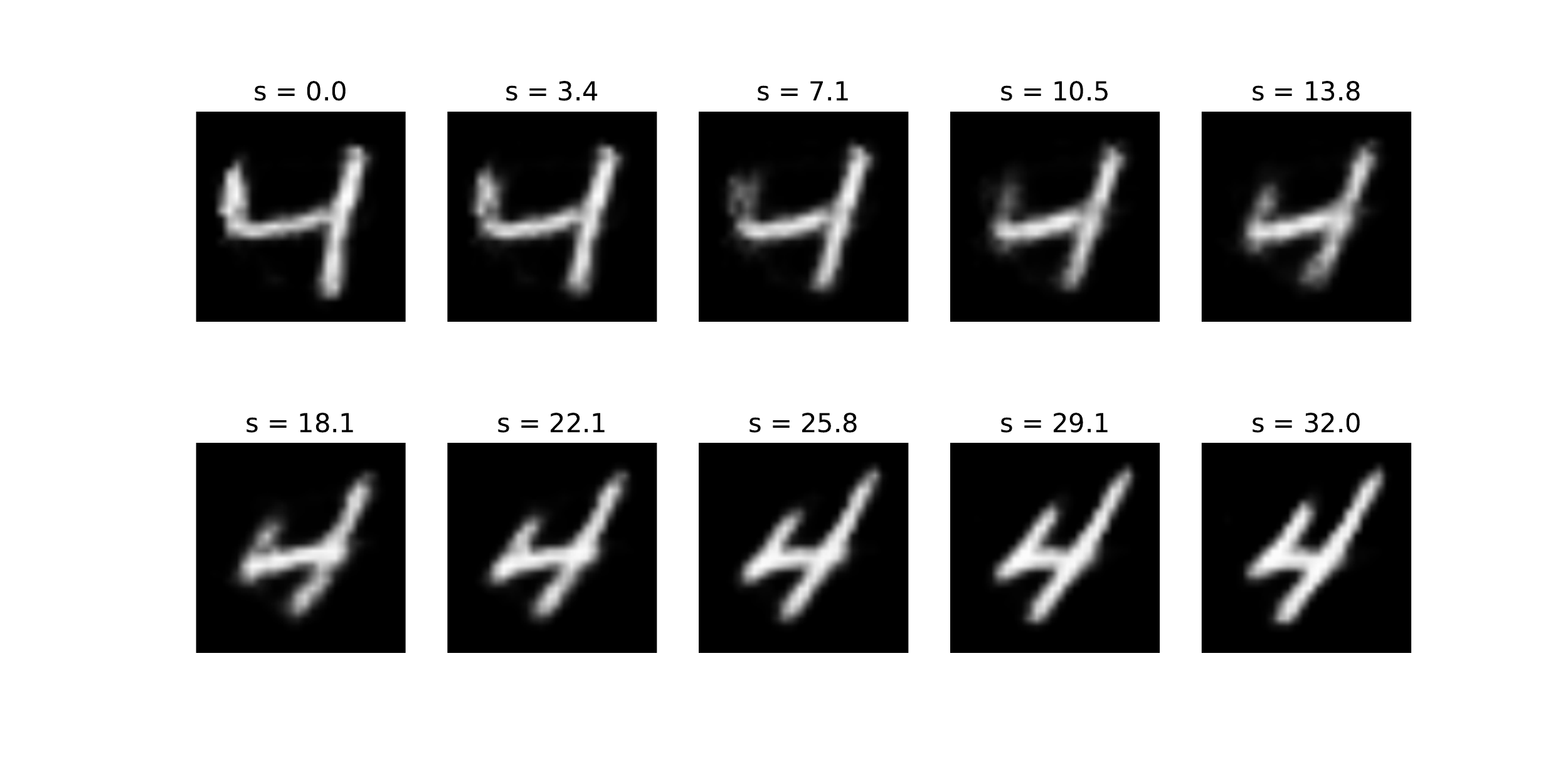} \qquad
    \includegraphics[scale=.275, trim=80 40 80 40, clip]{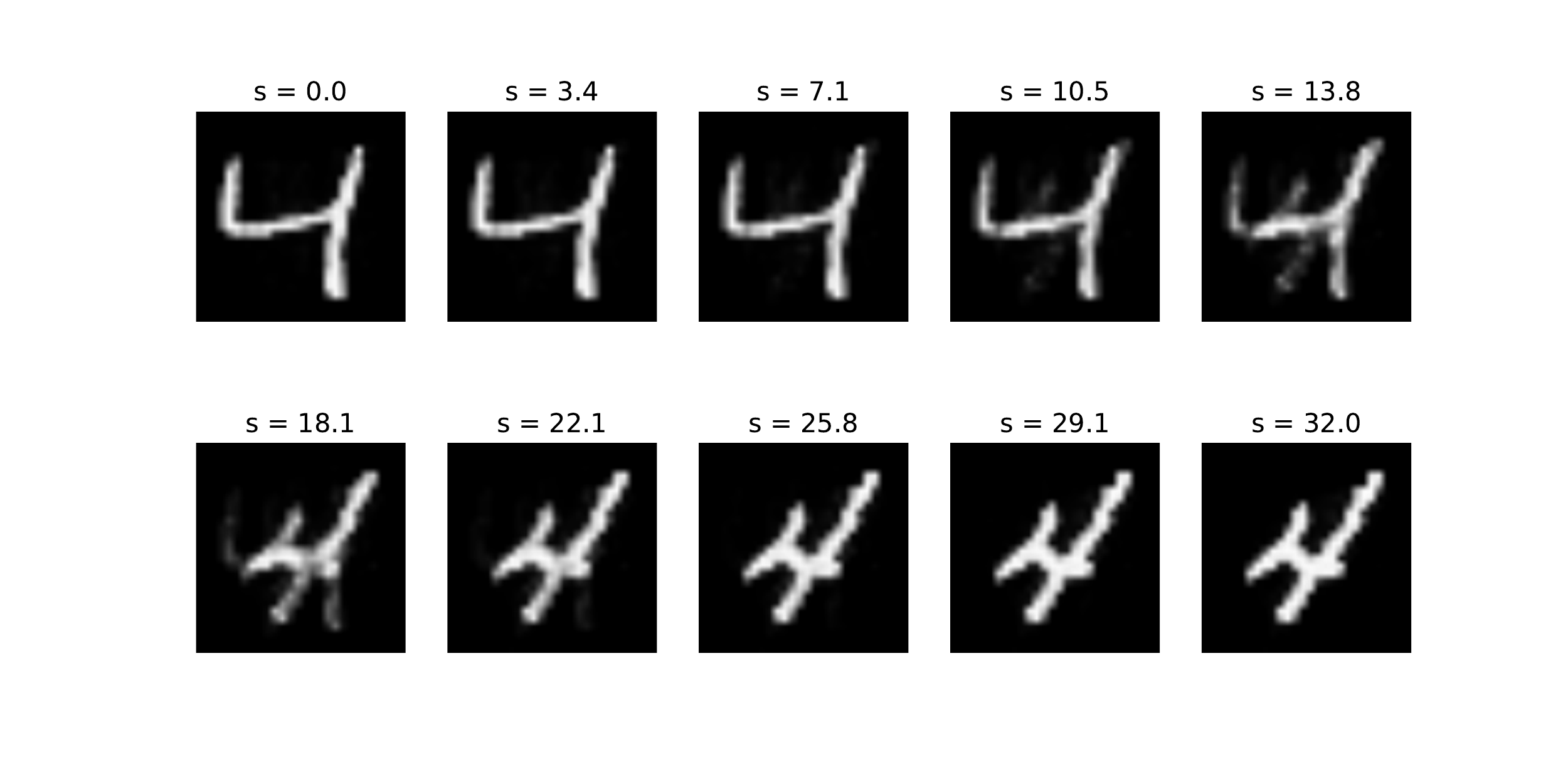}
    \includegraphics[scale=.4, trim=40 0 40 0, clip]{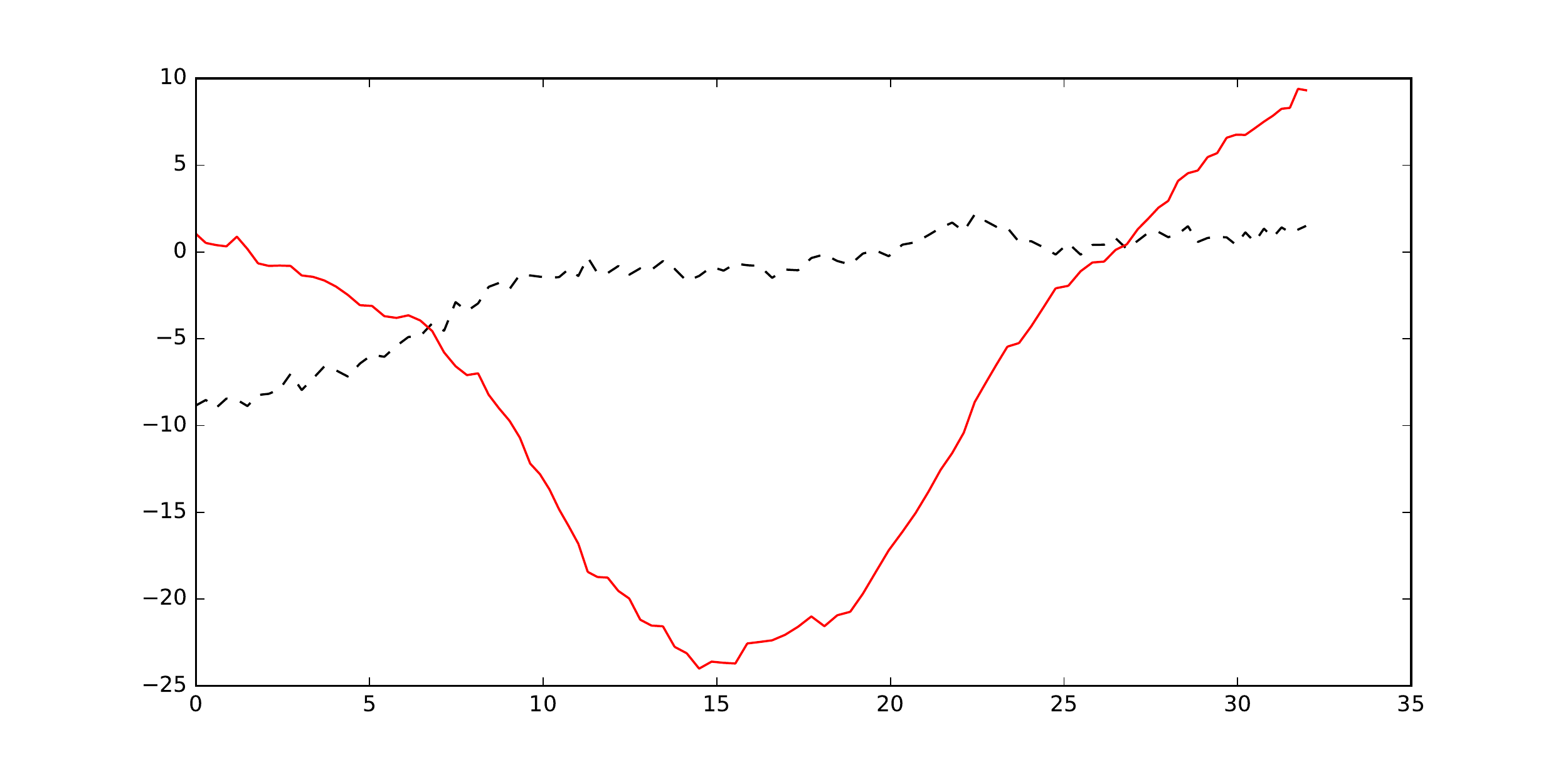}
    \end{center}
    
    \caption{Above left: a path in the well-trained network smoothly interpolates between two digits.  Above right: the same path in the overfitted network remains constant and then crossfades in the middle.  Below: $\log \widetilde p(s)$, i.e. log-probability as a function of arclength in $\X$.  Dashed is well-trained, solid is overfit.}
    
    \label{fig-fours}
\end{figure}

\begin{figure}[!htb]
    \begin{center}
    \includegraphics[scale=.275, trim=80 40 80 40, clip]{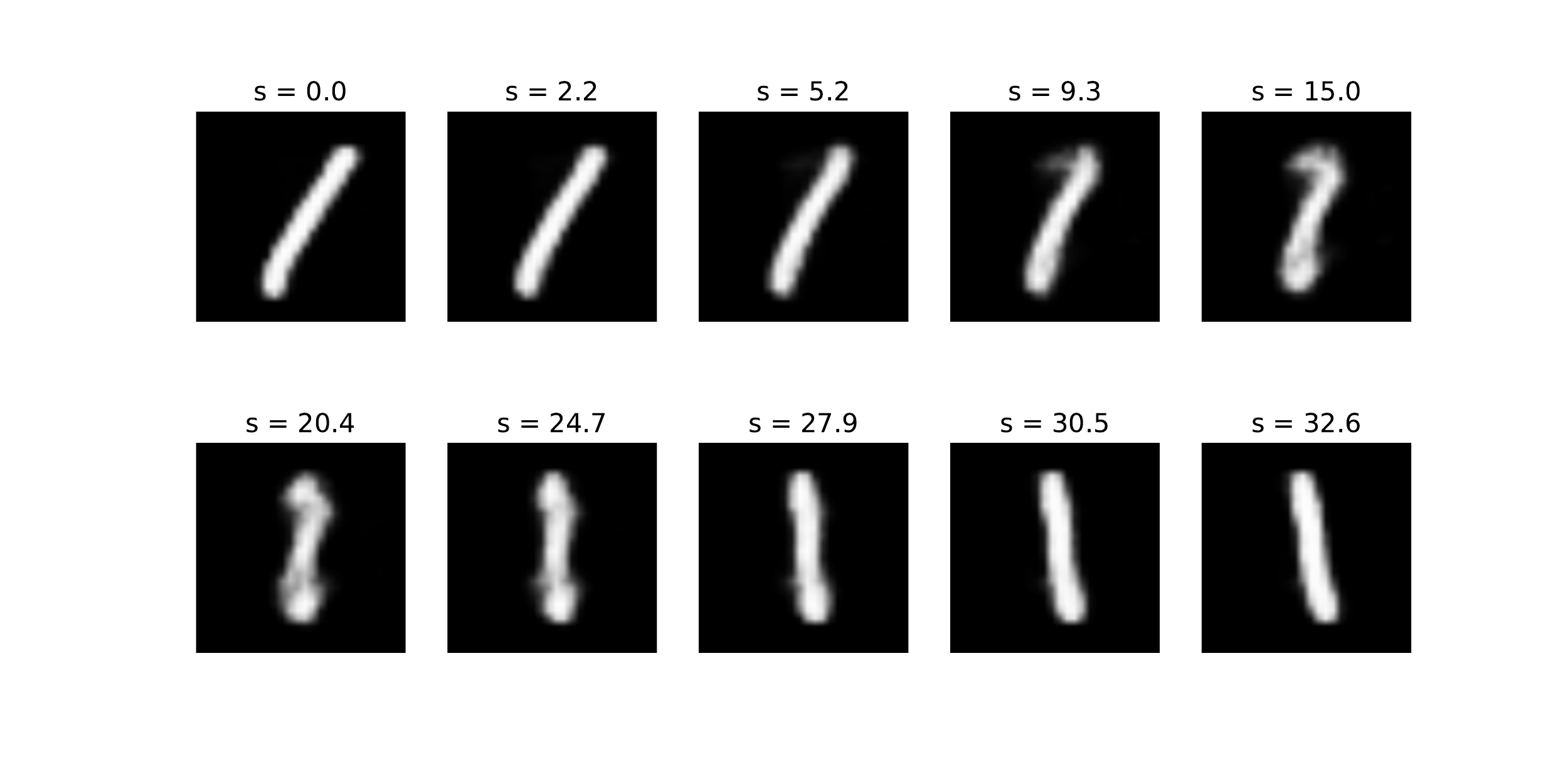} \qquad
    \includegraphics[scale=.275, trim=80 40 80 40, clip]{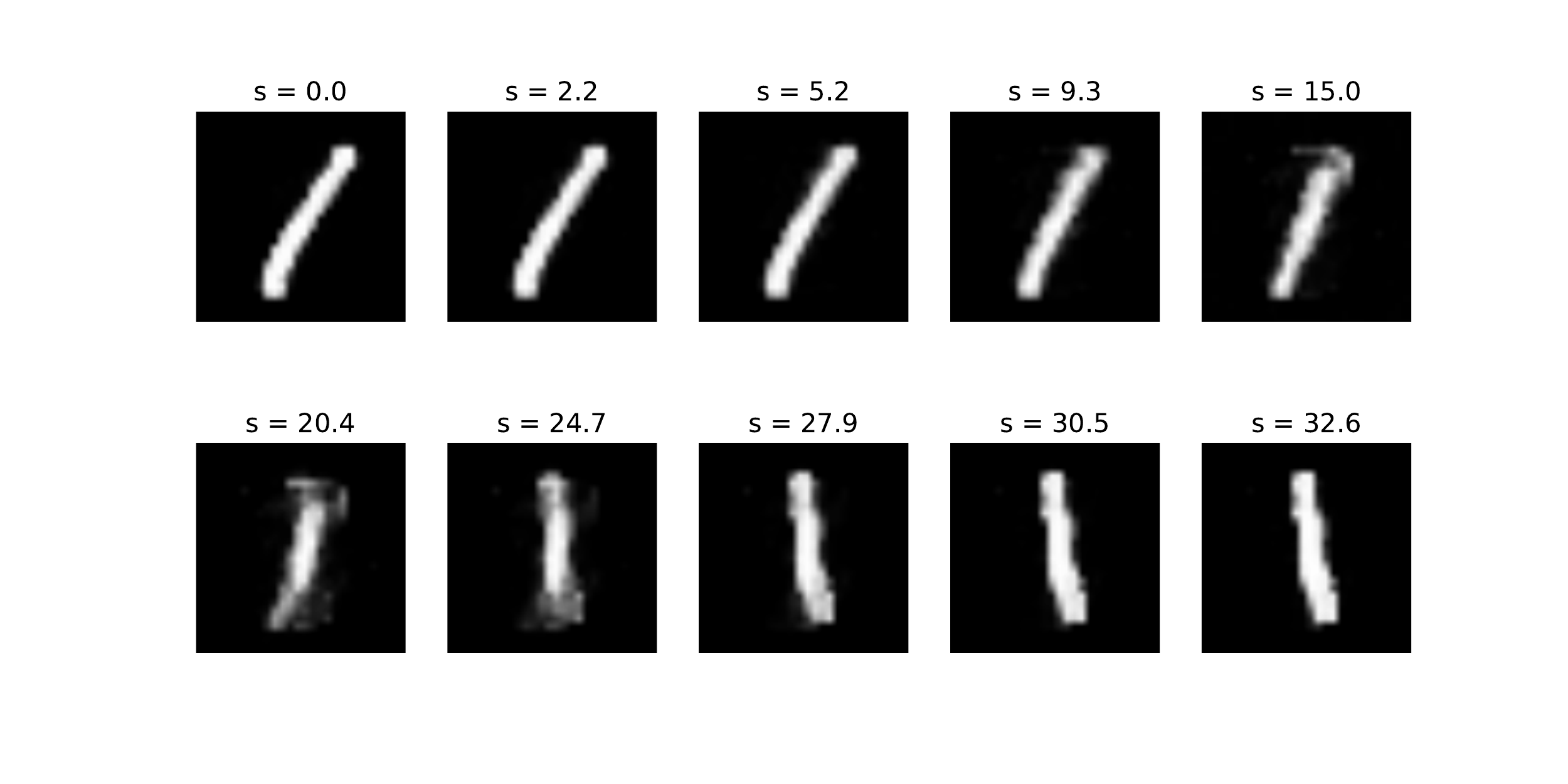}
    \includegraphics[scale=.4, trim=40 0 40 0, clip]{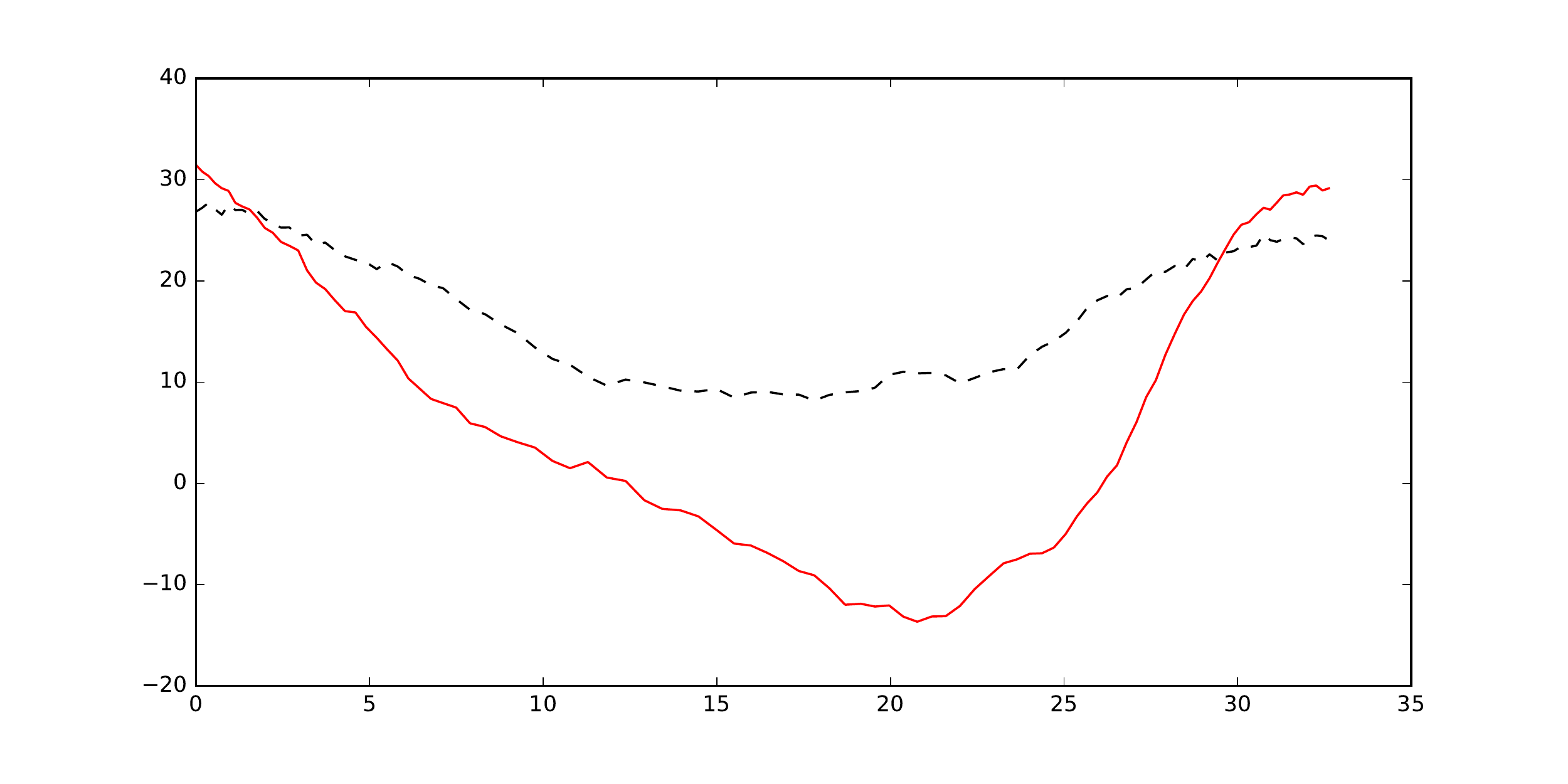}
    \end{center}
    
    \caption{Top left: well-trained.  Top right: overfit.  Bottom: $\log \widetilde p(s)$.  While the well-trained network interpolates better than the overfit network, the interpolation are visibly worse than figure \ref{fig-fours} and this is reflected in the dip in $\widetilde p$.  Dashed is well-trained, solid is overfit.}
    
    \label{fig-ones}
\end{figure}

Figures \ref{fig-fours}, \ref{fig-ones} show the two-point interpolation method of section \ref{section-lines}.  The endpoints of each path correspond to two training examples in the same class; the well-trained network does a much better job of interpolating the endpoints, and this is reflected in the log-density plots.  The well-trained network places significant mass in the regions between the endpoints, whereas the overtrained network places essentially all its mass at the endpoints.

\begin{figure}[!htb]
    \begin{center}
    \includegraphics[scale=.25, trim=60 700 60 40, clip]{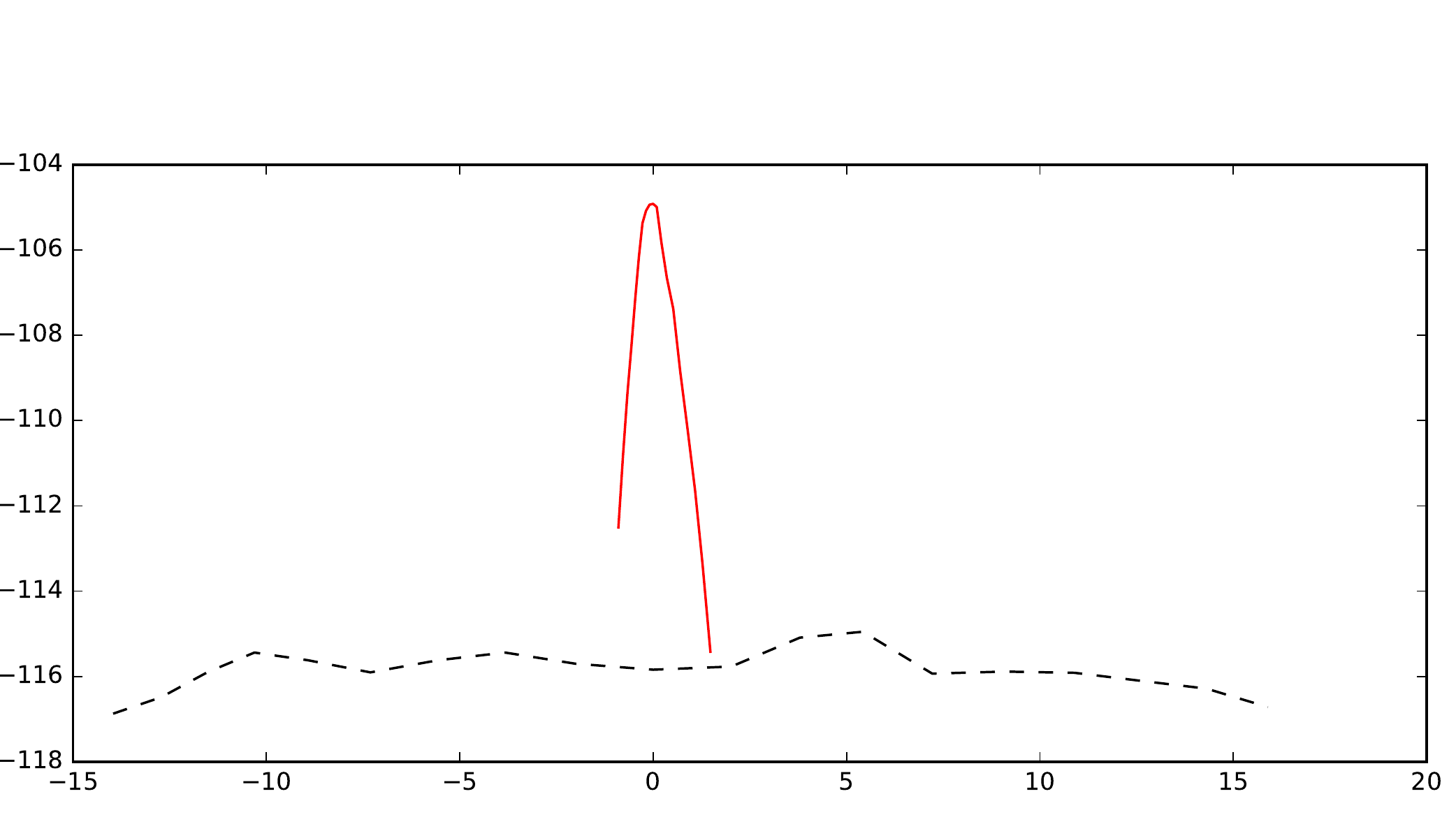} \qquad
    \includegraphics[scale=.25, trim=60 700 60 40, clip]{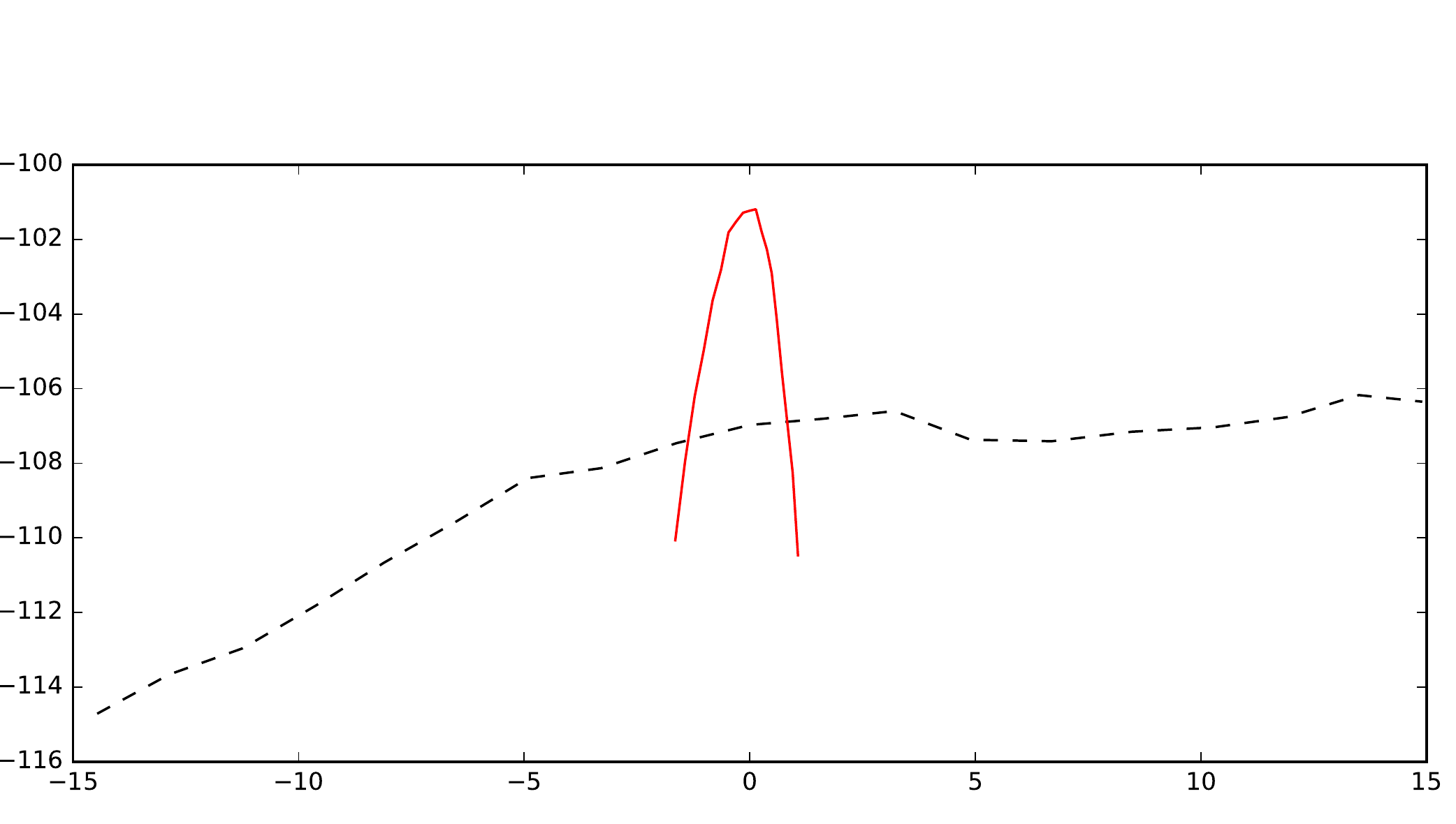}
    \end{center}
    
    \caption{Left: Decay of $\log \widetilde p(s)$ along the largest singular direction through a training sample of a "4".  Right: similar, through a "1".  Solid is overfit, dashed is well-trained.}
    
    \label{fig-fours-ones-decay}
\end{figure}

Figure \ref{fig-fours-ones-decay} shows the local decay method of section \ref{section-decay}.  These plots are obtained using a line in the largest singular direction, although similar plots are obtained using other non-degenerate singular directions.  The conclusion is similar: the overtrained network concentrates the density much more closely on each example.

\begin{figure}[!htb]
    \begin{center}
    \includegraphics[scale=.4, trim=40 700 40 40, clip]{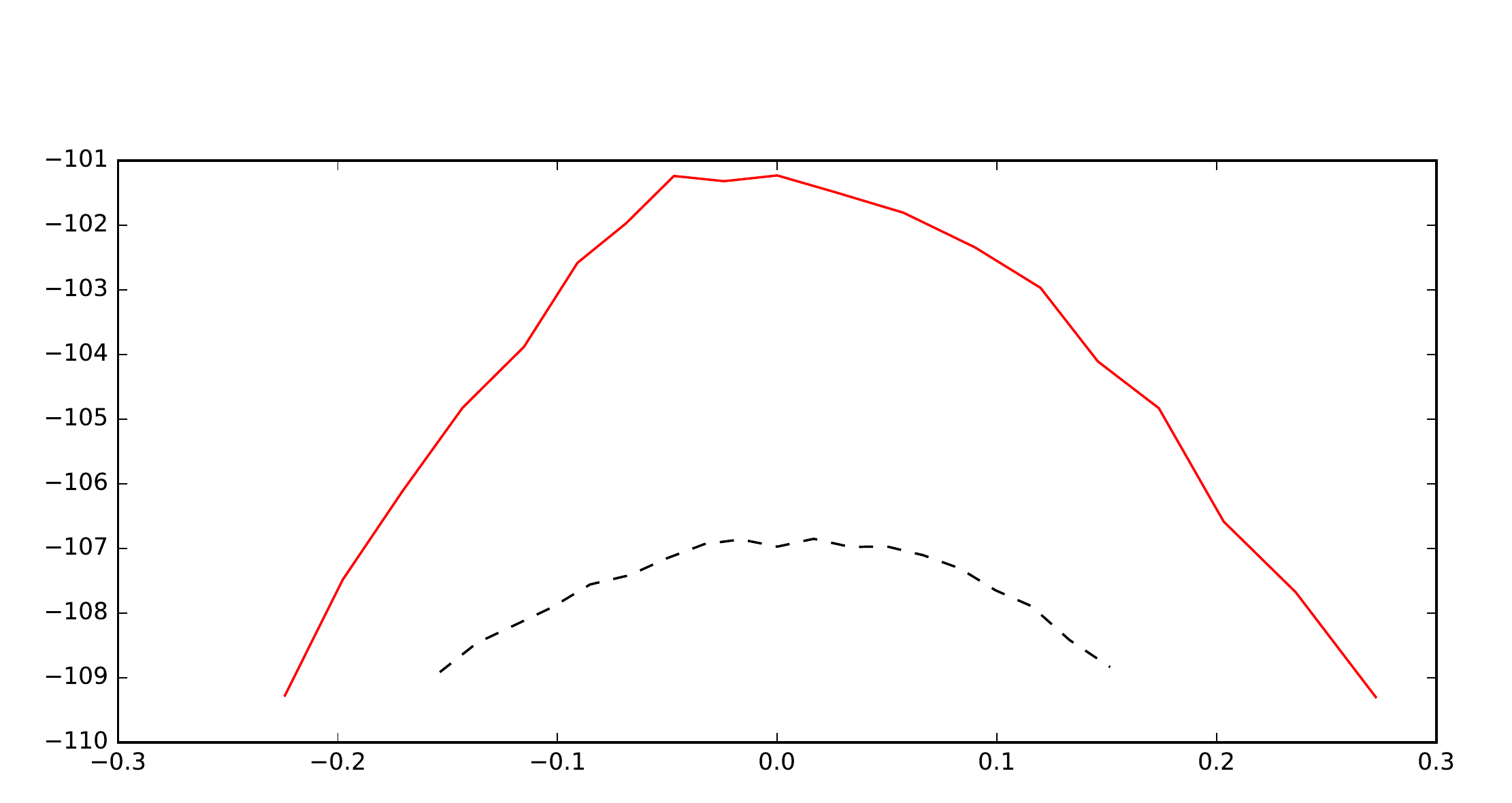}
    \end{center}
    
    \caption{Decay of $\log \widetilde p(s)$ in a direction with vanishingly small singular value, through a training sample of a "1".  Solid is overfit, dashed is well-trained.  Note the horizontal axis' scale: decay is almost immediate and both curves would essentially be delta functions if superimposed on figure \ref{fig-fours-ones-decay}.  (The greater absolute value of the overfit curve reflects the fact the the mass is concentrated on a handful of examples.) In any event, degenerate directions carry essentially no information and will only destabilize calculations.  }
    
    \label{fig-degenerate-direction}
\end{figure}

Finally, in figure \ref{fig-degenerate-direction} we show what happens when we look in degenerate directions: the plots become meaningless, as we're looking in directions in which $f$ is nearly constant.  Memorization and generalization become extremely difficult to distinguish based on the plots alone.

We obtain a single score from both methods as follows:
\begin{enumerate}
    \item We consider $\widetilde p(s)$ along lines joining each training sample to its nearest neighbor.  We compute second differences 
    \begin{align*}
        \delta_i = \log \widetilde p_i(s_0) + \log \widetilde p_i(s_1) - 2 * \log \widetilde p_i(s_{1/2})    
    \end{align*}
     $\widetilde p_i$ is the density along the $i$-th path, $s_0, s_1$ correspond to the endpoints of the path and $s_{1/2}$ is the midpoint.  We then average these second differences over the training set:
     \begin{align*}
         \mathrm{mean\ dip} = \frac{1}{N}\sum_{i=1}^N \delta_i
     \end{align*}
    
    \item For the local decay measure, we also consider second differences, but now at several radii about the central point along the largest singular direction, normalized by radius:
    \begin{align*}
        \eta_i(r) = \frac{1}{r}\left( \log \widetilde p_i(r) + \log \widetilde p_i(-r) - 2 * \log \widetilde p_i(0)\right)
    \end{align*}
    For a given set of radii $\{r_j\}_1^K$ we compute the mean decay of the peak:
    \begin{align*}
        \mathrm{mean\ decay} = \frac{1}{NK}\sum_{i=1}^N \sum_{k=1}^K \eta_i(r_k)
    \end{align*}
    The motivation for the multiscale method is to obtain robustness to various shapes of decay curves by averaging over several scales.  In experiments we use radii of $.5$ and $1$.
\end{enumerate}

Dip and peak results are summarized in the following table (note that positive dip and negative decay indicate memorization, since the curvatures are opposite):
\begin{center}
\begin{tabular}{ |c|c|c| } 
 \hline
 Model & Mean Dip & Mean Decay \\ 
 \hline
 Memorized & 1.07 & -3.40 \\ 
 \hline
 Well-trained & -0.242 & .0369 \\ 
 \hline
\end{tabular}
\end{center}

The results differ by an order of magnitude in each case.  The "Memorized" network exhibits a huge drop in log-probability in between samples and a very peaky density.  In contrast, the dip and peak scores for the well-trained network show that, if anything, the density increases slightly away from training samples.

\section{Conclusion}

"Memorization" in generative models means learning an output distribution which is concentrated on a finite number of output examples.  We have introduced methods for studying the output distribution and its concentration in the case where the latent density is easily to evaluate and the generator is a fixed function which is difficult to invert.  The main difficulty (the apparent degeneracy of the generator function $f: \R^m \to \R^n$) is overcome by noting that it is in fact a smooth map between submanifolds $\Z \subset \R^m$ and its image $\X \subset \R^n$ and we introduce machinery for computing the induced density on $\X$.

\section*{References}

\small




[1] Radford, A \ \& Metz, L \ \& Chintala, S \ (2015) Unsupervised Representation Learning with Deep Convolutional Generative Adversarial Networks {\it arXiv preprint} abs/1511.06434

[2] Gregor, K  \ \& Danihelka, I \ \& Graves, A \ \& Wierstra, D \ (2015) DRAW: A Recurrent Neural Network For Image Generation {\it arXiv preprint} abs/1502.04623

[3] Shifrin, T. \ (2005) {\it Multivariable mathematics : linear algebra, multivariable, calculus, and manifolds} Hoboken, NJ: Wiley

[4] do Carmo, M \ (2013) {\it Riemannian Geometry} Boston, MA: Birkhauser Boston
\end{document}